%% file: CNNJournal.tex
\pdfoutput=1

\documentclass[letterpaper, 10 pt, conference]{IEEEtran}  % Comment this line out if you need a4paper

\IEEEoverridecommandlockouts                              % This command is only needed if 
\usepackage{graphics} % for pdf, bitmapped graphics files
\usepackage{epsfig} % for postscript graphics files
\usepackage{mathptmx} % assumes new font selection scheme installed
\usepackage{times} % assumes new font selection scheme installed
\usepackage{amsmath} % assumes amsmath package installed
\usepackage{amssymb}  % assumes amsmath package installed

\usepackage[T1]{fontenc}
\usepackage{amsfonts}
\usepackage{booktabs}
\usepackage{siunitx}

\usepackage{booktabs}
\usepackage{graphicx}
\usepackage[ruled,lined]{algorithm2e}
\usepackage{float}
\usepackage[percent]{overpic}

\usepackage{mwe,tikz}\usepackage[percent]{overpic}

\usepackage{subcaption}
\usepackage{array,   tabularx, makecell, booktabs}%
\usepackage{geometry}
\usepackage{comment}

\title{\LARGE \bf
Obstacle-aware Waypoint Generation for Long-range Guidance of Deep-Reinforcement-Learning-based Navigation Approaches
}

\author{Linh K{\"a}stner$^{1}$\thanks{$^{1}$Linh K{\"a}stner, Xinlin Zhao, Zhengcheng Shen, and Jens Lambrecht are with the Chair Industry Grade Networks and Clouds, Faculty of Electrical Engineering, and Computer Science,				
		Berlin Institute of Technology, Berlin, Germany
		{\tt\small linhdoan@tu-berlin.de}}, Xinlin Zhao$^{1}$, Zhengcheng Shen$^{1}$ and Jens Lambrecht$^{1}$
}

\begin{document}

\maketitle
\thispagestyle{empty}
\pagestyle{empty}

\input{ieeeconf/content/0-abstract}
\input{ieeeconf/content/1-introduction}
% \newgeometry{top=0.75in,bottom=0.75in,right=0.75in,left=0.75in}
\input{ieeeconf/content/2-Related-Works}
\input{ieeeconf/content/3-methodology}

\input{ieeeconf/content/4-evaluations}

\input{ieeeconf/content/5-conclusion}

%\section*{Acknowledgement}
%We acknowledge help with the production of the spot welds and with the chisel test by Hubert Suwala.

\addtolength{\textheight}{-1cm} 

%%%%%%%%%%%%%%%%%%%%%%%%%%%%%%%%%%%%%%%%%%%%%%%%%%%%%%%%%%%%%%%%%%%%%%%%%%%%%%%%

%%%%%%%%%%%%%%%%%%%%%%%%%%%%%%%%%%%%%%%%%%%%%%%%%%%%%%%%%%%%%%%%%%%%%%%%%%%%%%%%

%%%%%%%%%%%%%%%%%%%%%%%%%%%%%%%%%%%%%%%%%%%%%%%%%%%%%%%%%%%%%%%%%%%%%%%%%%%%%%%%

%%%%%%%%%%%%%%%%%%%%%%%%%%%%%%%%%%%%%%%%%%%%%%%%%%%%%%%%%%%%%%%%%%%%%%%%%%%%%%%%
\typeout{}
\bibliographystyle{IEEEtran}
\bibliography{ref}

\end{document}

%% file: ieeeconf/content/0-abstract.tex
% !TeX encoding = utf-8
% !TeX language = en_GB
% !TeX spellcheck = en_GB
% !TeX root = paper.tex

\begin{abstract}

Navigation of mobile robots within crowded environments is an essential task in various use cases, such as delivery, health care, or logistics. Deep Reinforcement Learning (DRL) emerged as an alternative method to replace overly conservative approaches and promises more efficient and flexible navigation. However, Deep Reinforcement Learning is limited to local navigation due to its myopic nature. Previous research works proposed various ways to combine Deep Reinforcement Learning with conventional methods but a common problem is the complexity of highly dynamic environments due to the unpredictability of humans and other objects within the environment. In this paper, we propose a hierarchical waypoint generator, which considers moving obstacles and thus generates safer and more robust waypoints for Deep-Reinforcement-Learning-based local planners. Therefore, we utilize Delaunay Triangulation to encode obstacles and incorporate an extended hybrid A-Star approach to efficiently search for an optimal solution in the time-state space. We compared our waypoint generator against two baseline approaches and outperform them in terms of safety, efficiency, and robustness.

\end{abstract}

%% file: ieeeconf/content/1-introduction.tex
\section{Introduction}
Autonomous robot navigation is a fundamental aspect in robotics, especially for service robots, logistic robots, and autonomous vehicles \cite{fragapane2020increasing}. While navigation in a known static environment is a well-studied problem, navigation in a dynamically changing environment remains a challenging task for many mobile robot applications. It requires the robot to efficiently generate safe actions in proximity to unpredictably moving obstacles in order to avoid collisions. Traditional model-based motion planning approaches employ hand-engineered safety rules to avoid dynamic obstacles. However, hand-designing the navigation behavior in dense environments is difficult since the future motion of the obstacles is unpredictable \cite{qian2010socially}. Deep Reinforcement Learning (DRL) emerged as a learning-based method that can learn policies by interacting with the environment on a trial-and-error basis. Due to the ability to learn nonlinear patterns, DRL is employed to enable the robot to learn the complex behavior rules, environment semantics, and add common-sense reasoning into navigation policies. However, DRL tends to be extremely sample-inefficient and highly specialized to the system they were trained on \cite{recht2019tour}. Moreover, due to the myopic nature of DRL, a variety of literature incorporates DRL into robot navigation systems only as a local planner, where the sample space is smaller and the planning horizon is restricted locally \cite{dugas2020navrep}.
To tackle these limitations of DRL and leverage its advantages for navigation in long-range dynamic environments, long-term global guidance such as the provision of a subgoal is required to relieve its myopic nature. However, real-time generation of a globally consistent path in long-range navigation tends to be computationally expensive. Moreover, most global planning approaches are based on static environment assumptions \cite{demyen2006efficient} \cite{harabor2019regarding} \cite{moon2014kinodynamic}. Unlike in a static environment, in a dynamic environment, the planned global path at an arbitrary point in time could be immediately invalid for the next moment due to dynamic obstacles.
On that account, we direct our attention to the challenges associated with navigation in a long-range, dynamic environments and the idea of effective and efficient exploration in the state-time space of these environments by using traditional model-based methods to provide global guidance to a DRL-based local planner in real -time.
The main contributions are as follows:
\begin{itemize}
    \item Proposal of a global planner based on Hybrid A-Star, which generates landmark waypoints as sparse representation of the global path.

    \item Proposal of an efficient trajectory planning method for the mid planner, which integrates a front-end searching algorithm based on timed Delaunay triangle graph for planning a near-optimal initial path and a back-end ESDF-free gradient-based trajectory optimization algorithm. The trajectory takes moving obstacles into consideration and provides mid-term global guidance (subgoal) in a highly dynamic environment.
    
    \item Evaluation of the navigation system against two baseline approaches in highly dynamic environments in terms of safety robustness and efficiency against conventional planning methods.
\end{itemize}
The paper is structured as follows: Sec. II begins with related works, followed by the methodology in Sec. III. Subsequently, the results and evaluations are presented in Sec IV. Finally, Sec. V provides a conclusion and outlook.
We made our code open-source at https://github.com/ignc-research/arena-fsm-ego-planner.

%% file: ieeeconf/content/2-Related-Works.tex
\section{Related Works}
Among the most common traditional OA approaches are reactive methods such as velocity obstacles \cite{van2008reciprocal}, \cite{van2011reciprocal}, artificial potential field \cite{park2001obstacle}, \cite{sun2017collision} or vector field histograms \cite{borenstein1989real}. However, these methods require high computational calculations and can not cope well with fast moving obstacles. Other traditional obstacle avoidance approaches contain the model predictive control (MPC) \cite{rosmann2019time}, timed elastic bands (TEB) \cite{rosmann2015timed} or the dynamic window approach (DWA) \cite{fox1997dynamic}, which repeatedly solve optimization control problems.
DRL-based navigation approaches proved to be a promising alternative that has been successfully applied in various robotic applications with remarkable results. Various works demonstrated the superiority of DRL-based OA approaches due to more flexiblility in the handling of obstacles, generalization new problem instances, and ability to learn more complex tasks without manually designing the functionality \cite{faust2018prm}, \cite{everett2018motion}, \cite{chen2019crowd}.
However, since the reward that a DRL agent can obtain in long-range navigation over large-scale maps is usually sparse, agents are only suitable for short-range navigation due to local minima. Thus, a variety of research works combine DRL-based local planning with traditional methods through the use of waypoints to provide the DRL approach with short-range goals on a global path generated by a traditional global planner such as RRT \cite{lavalle1998rapidly} or A-Star \cite{hart1968formal}.
Gundelring et al. \cite{guldenringlearning} first integrated a DRL-based local planner with a conventional global planner from the ubiquitously used robot operating system (ROS) and demonstrated promising results. The researchers employ a subsampling of the global path to create waypoints for the DRL-local planner.
Similarly, Regler et al. \cite{regier2020deep} propose a hand-designed sub-sampling to deploy a DRL-based local planner with conventional navigation stacks.
A limitation of these works is that the simple sub-sampling of the global path is inflexible and could lead to hindrance in complex situations, e.g. when multiple humans are blocking the way. Other works employed a more intelligent way to generate waypoints.
Brito et al. \cite{brito2021go} proposed a DRL-based waypoint generation where the agent is trained to learn a cost-to-go model to directly generate subgoals, which an MPC planner follows. The better estimated cost-to-go value enables MPC to solve a long-term optimal trajectory. Similarly, Bansal et al. \cite{bansal2020combining} proposed a method called LB-WayPtNav, in which a supervised learning-based perception module is used to process RGB image data and output a waypoint. With the waypoint and robot current state, a spline-based smooth trajectory is generated and tracked by a traditional model-based, linear feedback controller to navigate to the waypoint. However, training of DRL-based agents is complex and not always intuitive requiring a lot of pre assumptions and limitations. Supervised training, requires a tedious data acquisition phase to provide annotated training data.
In this work, we follow a similar hierarchical approach but focus on employing a model-based waypoint generator combined with a DRL-based local planner. This way, we leverage the superior performance of DRL-based approaches for the obstacle avoidance problem.
Inspired by recent works of Zhou et al. \cite{zhou2020ego}, which efficiently search the state-time space for an optimal trajectory considering fast-moving obstacles, our waypoint generator is able to provide meaningful waypoints without prior training while considering fast-moving obstacles.

%% file: ieeeconf/content/3-methodology.tex
\begin{figure*}[!h]
    \centering
		\includegraphics[width= 0.99\textwidth]{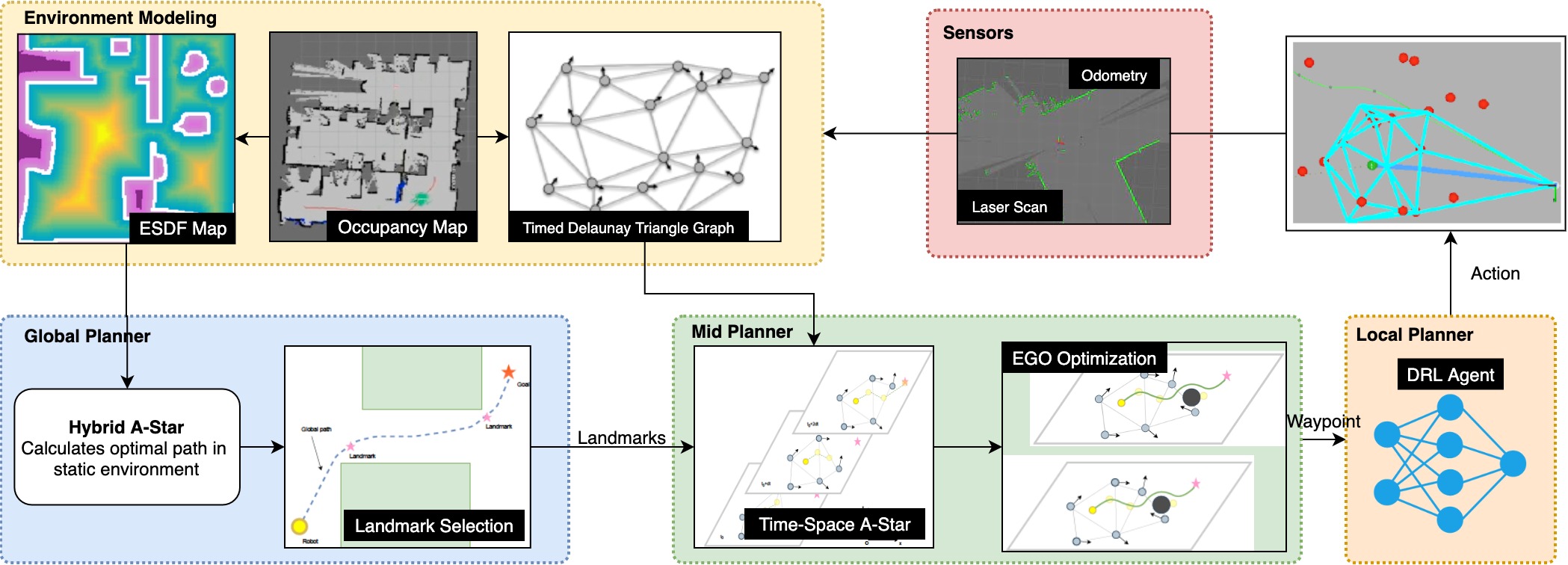}
	\caption{System design. The hierachical navigation systems contains a variety of modules. Three modules are designed for different planning horizons. Different environment representations are also part of the system.}
	\label{system}
\end{figure*}

\section{Methodology}
In this section, we will present the methodology of our proposed framework. Our main objective is to explore and exploit a feasible and sub-optimal trajectory in state-time space effectively and efficiently to provide meaningful waypoints for a DRL-based local planner.

\subsection{System Design}
An overview of the complete system structure is shown in Fig. \ref{system}. The navigation system contains different layers and modules. A hierarchical motion planning framework is adopted to handle complex long-range navigation problems. Three modules are designed for different planning horizons. The global planner is designed to provide a global path to guide the robot and avoid local minima. Therefore, a hybrid A-Star search is implemented to generate a kinodynamic feasible global path. The mid planner is designed to generate a collision-free trajectory in the mid horizon that incorporates dynamic obstacles within the sensor range. The trajectory generates a subgoal/waypoint, which is given as input to the local planner.
Finally, the local planner is designed to plan the local horizon with the given waypoint from the mid planner. In the following section, each module is described in more detail.

\subsection{Environment Representation}
A reasonable representation of the environment is as important as the motion planning algorithms themselves. An appropriate environment representation can improve the effectiveness and efficiency of these processes and thus speed up the motion planning process. In this work, we make use of three different representations: the occupancy grid map, the Euclidean signed distance field (ESDF) map, and the Delaunay triangle graph, which is modified to cope with dynamic obstacles.

\subsubsection{Occupancy Grid Map}
Typically, planning approaches use occupancy grid maps that indicate whether a pixel is occupied, not occupied, or unknown. Although this representation is efficient and simple, it does not contain enough information about obstacles that could be of relevance. Thus, we additionally rely on other map representations that provide more information about the obstacles.

\subsubsection{ESDF Maps}
The ESDF map gives the minimum distance of each cell to the obstacles around and thus provides relevant information for collision checking approaches.
It is built based on the Fast Incremental Euclidean Distance Fields approach for Online Motion Planning of Aerial Robots (FIESTA) \cite{han2019fiesta}, which incrementally builds an ESDF map from the occupancy grid directly.
In this work, we utilize the ESDF map for global path optimization and update it once at the initialization of the system.

\subsubsection{Delaunay Triangulation}

The Delaunay triangle graph is a sparse topological map and is used to represent topological relations between dynamic obstacles, which is the key aspect to speed up the exploration process in the dynamic state-time space.
With the observation that the dynamic obstacles such as other robot agents and pedestrians are usually spread sparsely in the environment while each of them occupying only a small area in the space, we can model the dynamic obstacles with a rather sparse map and to find a collision-free trajectory from the gaps between the obstacles. Delaunay triangulation is an intuitive representation of the dynamic environment, in which the position of dynamic obstacles can be seen as the vertices of triangle. 

\subsection{Global Planner}
Our proposed global planner module consists of two phases: first, a collision-free path is calculated using a hybrid A-Star search. Second, a landmark generation module computes suitable points along the global path.

\subsubsection{Hybrid A-Star search}
The global planner incorporates a hybrid A-Star search to generate a kinodynamic feasible global path given an occupancy and an ESDF map. Hybrid A-Star extends the clasic A-Star search by searching directly in the state space for a collision-free and kinodynamic feasible trajectory while minimizing the time duration and control cost. The key difference from standard A-Star is that the edges connecting two nodes are not straight-line segments but motion primitives, which are continuous local trajectories integrated by samples in the action space. To limit the growth of the search graph in the exploration space, the discretized grid is used, while nodes can reach any continuous point on the grid. In this paper, a simplified Hybrid A-Star is used and in order to accelerate the global planning process, we simplify the original Hybrid A-Star in two aspects: the trajectory dimension is reduced to 2D with x-axis and y-axis and the kinematic model is reduced to a double integrator.

\subsubsection{Landmark Generation}
\begin{figure}[!h]
    \centering
        \includegraphics[width=0.5\textwidth]{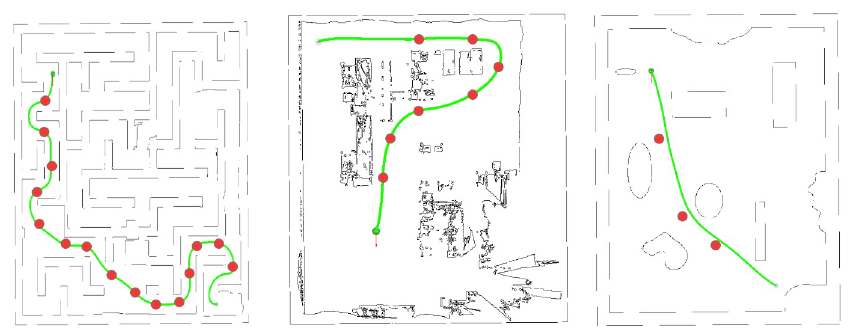}
    \caption{Exemplary process of the timed A-Star search}
    \label{landmarks}
\end{figure}

Since not all points on the global path are equally important, only a few critical points are used. As a result, the search space can be limited, and only reasonable waypoints proposed for the mid planner. These points are denoted as 'Landmark waypoints', which are defined as critical waypoints that the robot must pass in order to reach the goal, such as turning points at corners. Exemplary landmarks on a calculated global path are illustrated in Fig. \ref{landmarks}.

These landmark waypoints are a sparse representation of the global path, which is advantageous in dynamic environments. By traversing each landmark waypoint sequentially, the robot is able to reach the goal position without trapping by local minima. A landmark waypoint is defined as a point at a position $[p_x,p_y]$, where the required steering angle $\delta$ to traverse this point is larger than a threshold value $\delta_{th}$.
 
 \begin{equation}
     \delta > \delta_{th}
     \label{equ:steering_angle}
 \end{equation}

As for a differential drive robot, the heading angle of the robot $\theta$ is controlled directly by the difference in wheel velocities. Therefore, the change of the heading angle $\Delta \theta$ within an amount of time is used to reflect the steering angle $\delta$ of the robot. Thus, the criterion for selecting landmark waypoints is transformed to finding the points, where the change of heading angle within a time interval is larger than a threshold value $\Delta \theta > \Delta \theta_{th}$. Since the global time-parameterized trajectory is obtained by performing Hybrid A-Star search, the angular velocity trajectory $\omega(t)$ of the heading angle can be computed by following kinematic relation:

\begin{align}
    \mathbf{a}_{\text {norm}}(t)&=\dfrac{\mathbf{v}(t) \times \mathbf{a}(t)}{\|\mathbf{v}(t)\|} \\
    \omega(t)&=\dfrac{\left\|\mathbf{a}_{\text {norm }}(t)\right\|}{\|\mathbf{v}(t)\|}
\end{align}

where $\mathbf{v}(t)=\dot{\mathbf{p}}(t)$ is the velocity and $\mathbf{a}(t)=\ddot{\mathbf{p}}(t)$ is the acceleration in the 2D configuration space.
By integrating the angular velocity from a time interval $[t_0, t_i]$, the change of heading angle $\Delta\theta$ can be calculated using Equ. \ref{equ:delta_theata}
\begin{equation}
\Delta\theta = \int_{t_0}^{t_i} \omega(\tau) d \tau
\label{equ:delta_theata}
\end{equation}

The final algorithm of the landmark generation is formalized in Alg. \ref{alg:landmark} and the resultant global path and landmark waypoints are shown in Fig. \ref{landmarks}

\begin{algorithm}[]

  \caption{Landmark generation($\mathbf{x}_{start},\mathbf{x}_{goal}$)}

  \label{alg:landmark}
 $L \leftarrow \emptyset$\;
$\boldsymbol{\xi}_{global}(t)$ $\leftarrow$
 \textbf{Hybrid A-Star}($\mathbf{x}_{start},\mathbf{x}_{goal}$)\;
 $\Delta\theta \leftarrow 0$\;
 $l_{prev} \leftarrow \mathbf{p}_{start}$\;
  \For{$t \leftarrow t_s$ to $t_e$}{
    $\omega(t) \leftarrow \mathbf{ComputeAngularVelocity}(\mathbf{v}(t),\mathbf{a}(t))$\;
    $\Delta\theta \leftarrow \Delta\theta + \omega(t)\Delta{t}$\;
    \If{$\Delta\theta > \Delta\theta_{th} \wedge \mathbf{ComputeDist}({l_{prev}}, \mathbf{p}(t))> D_{th}$ }{
  ${L}.\mathbf{add}\left(\mathbf{p}(t)\right)$\;
       $l_{prev} \leftarrow \mathbf{p}(t)$\;
       $\Delta\theta \leftarrow 0$\;}}
 ${L}.\mathbf{add}(\mathbf{p}_{goal})$
 \end{algorithm}

\subsection{Mid Planner}
The mid planner consists of two modules, which we denote as front-end and back-end. The front-end takes as input the computed landmarks from the previously presented global planner and computes a trajectory considering dynamic obstacles based on Delaunay triangulation.
The back-end incorporates an ESDF-free gradient-based local trajectory optimizer (EGO) from \cite{zhou2020ego} et al. to optimize the trajectory to a collision-free dynamic feasible trajectory in real-time. The subgoal is selected on the final optimized trajectory.
In the following, each part is described in more detail.

\subsubsection{Frontend - State-Timed A-Star-search}
For the front-end, a state-time A-Star search based on a timed Delaunay triangle graph is designed to find an initial trajectory considering the goal and obstacles information.
It provides an initial trajectory to the back-end to be further optimized by the EGO trajectory optimizer. The state-time A-Star search is an extension of standard A-Star based on a sparse representation of state-time space called the timed Delaunay triangle graph. The resultant trajectory takes both dynamic obstacles in the local sensor range and the input landmark waypoint as a mid-term goal into consideration, thus it can provide better global guidance to local DRL planners.

\begin{figure}[!h]
    \centering
        \includegraphics[width=0.28\textwidth]{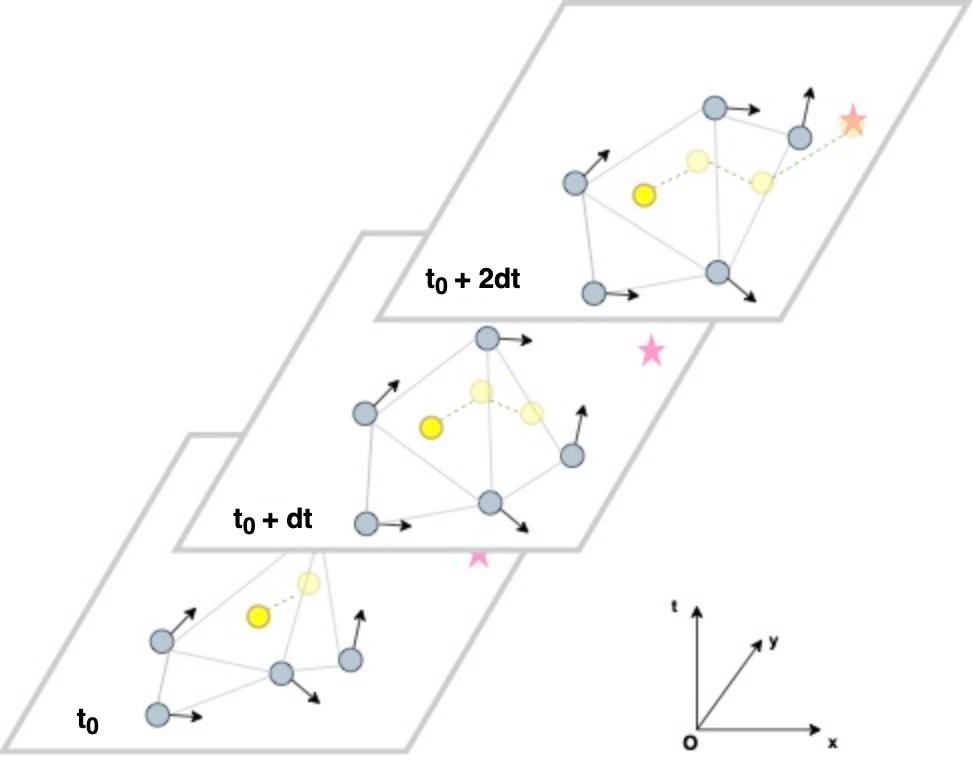}
    \caption{Exemplary process of the timed A-Star search}
    \label{fig:state_time_graph}
\end{figure}

The approach utilizes a Delaunay triangle graph $G[t]$, which encodes dynamic obstacles as vertices of the triangles. Due to the sparsity of the topological representation, the exploration phase can be done efficiently. The uncertainty of future motions of dynamic obstacles is relieved by rapid iterative replanning. Thus, the error caused by constant velocity assumptions in predicting obstacle movements can be reduced. An example of the search process of state-time A-Star is shown in Fig. \ref{fig:state_time_graph}.
At the beginning of each planning cycle, a discretized timed Delaunay triangle graph containing a set of triangle graphs is constructed according to the start state $\mathbf{x}_{start}$, the goal state $\mathbf{x}_{goal}$ and current perceived states of dynamic obstacles $O[t_0]$ within the sensor range. The timed graph G contains a set of Delaunay triangle graphs from the time $t_0$ to $t_0+T_h$, where $T_h$ is the planning time horizon. The state-time space exploration is done by generating a simplified trajectory through the edge connecting two states. Then, the samples in the sample set $ST$ are pruned so that only the sample that stays safe from the collision within the planning time horizon are considered as neighbors. Subsequently, not only are the \textit{time-to-collision} conditions of each sample checked, but a re-computation for a new safe sample based on the original sample is performed if the original sample is expected to collide with obstacles during the action duration. This sampling mechanism is inspired by RRT and ensures the algorithm to have enough effective samples in the state-time space. After valid neighbor nodes are found, an update of the sample set $ST$ is performed according to $g$-cost and heuristics. The heuristic calculation is designed to take the time-to-avoid dynamic obstacles into consideration. Since the sampling algorithm is based on the triangle graph, which contains the goal state (here the landmark waypoint) as a vertex of the triangle, the proposed state-time A-Star method is able to incorporate global information as well as the dynamic local environment within the sensor range simultaneously.

\subsubsection{Backend- EGO Optimization}
For the last module in our pipeline, we incorporate an ESDF-free gradient-based trajectory optimizer (EGO) by Zhou et al. \cite{zhou2020ego} as the back-end of the mid planner to optimize the initial trajectory computed by the state-time A-Star search.
\begin{figure}[!h]
    \centering
        \begin{subfigure}{0.28\textwidth}(a)
         \centering % <-- added
        \includegraphics[width=\linewidth]{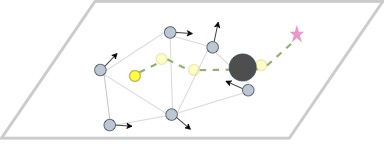}
        
        \label{fig:1}
    \end{subfigure}\hfil % <-- added
    \begin{subfigure}{0.28\textwidth }(b)
     \centering % <-- added
        \includegraphics[width=\linewidth]{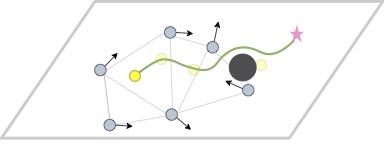}
        
        \label{fig:2}
    \end{subfigure}
    \caption{Comparison between the initial trajectory (a) and the optimized trajectory (b). The dark grey circle is an unexpected obstacle not considered by the initial state-time A-Star search, the green dashed line is the initial trajectory, the green line is the optimized trajectory.}
    \label{fig:optimization_compare}
\end{figure}

Although the dynamics of the robot and obstacle are considered in state-time A-Star search, due to simple constant linear assumption of dynamic obstacles motion, the generated trajectory is not smooth and cannot ensure the real-time safety of the robot. However, this trajectory is a reasonable initial guess, which has considered long-term information of the global landmark waypoint and local dynamic obstacles. The incorporation of the EGO trajectory optimizer should further enhance navigational safety.
The key advantage of the EGO optimizer is that it provides fast real-time optimization. Moreover, unlike other gradient-based trajectory optimization methods, the EGO optimizer can also deal with an initial trajectory that is in a collision and optimize it to a collision-free dynamic feasible trajectory as shown in Fig. \ref{fig:optimization_compare}
The trajectory optimization of the EGO optimizer requires three steps: trajectory parameterization of the initial trajectory into a B-spline, artificial distance field generation for collision optimization, and the numerical optimization process of the trajectory given optimization objective functions. For a detailed explanation, we refer to \cite{zhou2020ego}.

%% file: ieeeconf/content/4-evaluations.tex
\begin{figure*}[!h]
	
		\begin{subfigure}{0.31\textwidth}(i)
		 \centering % <-- added
		\includegraphics[width=\linewidth]{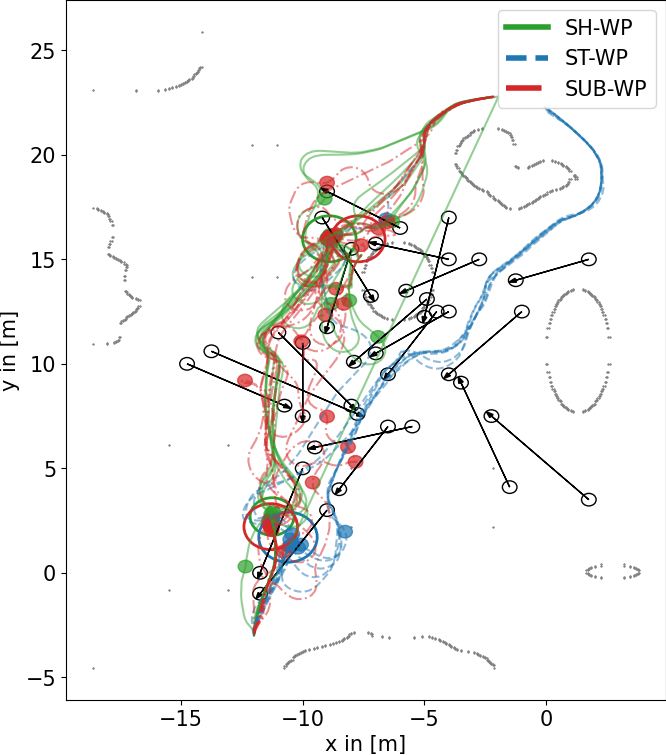}
		
		\label{fig:1}
	\end{subfigure}\hfil % <-- added
	\begin{subfigure}{0.31\textwidth }(ii)
	 \centering % <-- added
		\includegraphics[width=\linewidth]{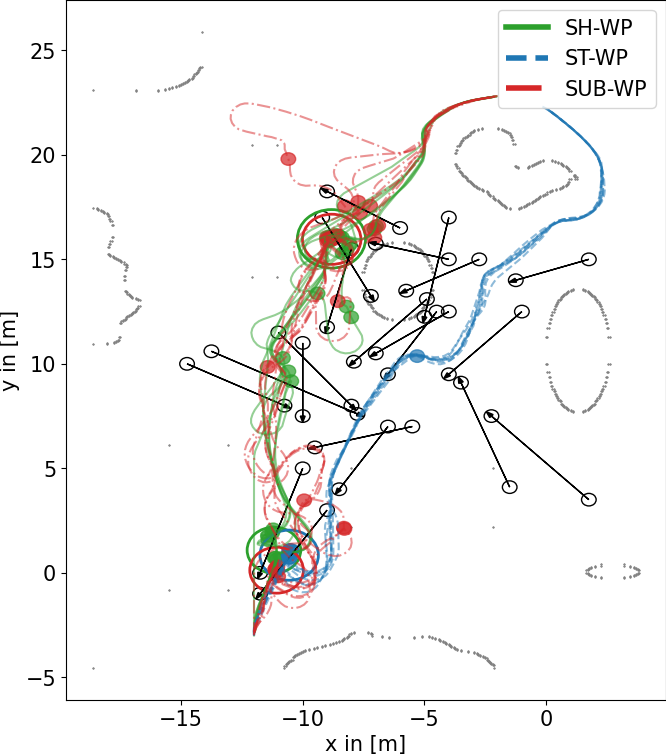}
		
		\label{fig:2}
	\end{subfigure}\hfil % <-- added
	\begin{subfigure}{0.31\textwidth}(iii)
	 \centering % <-- added
		\includegraphics[width=\linewidth]{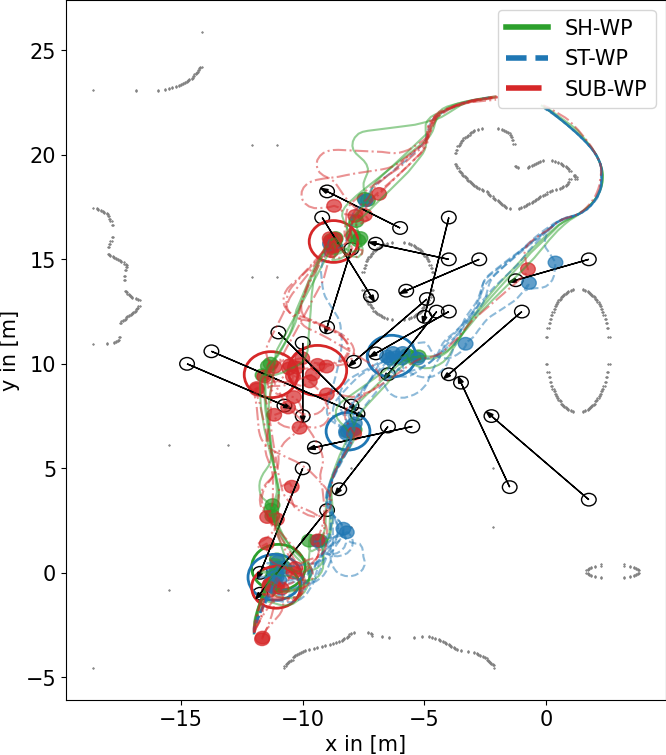}
		
		\label{fig:3}
	\end{subfigure}
	\medskip
	\begin{subfigure}{0.25\textwidth}(a)
		 \centering % <-- added
		\includegraphics[width=\linewidth]{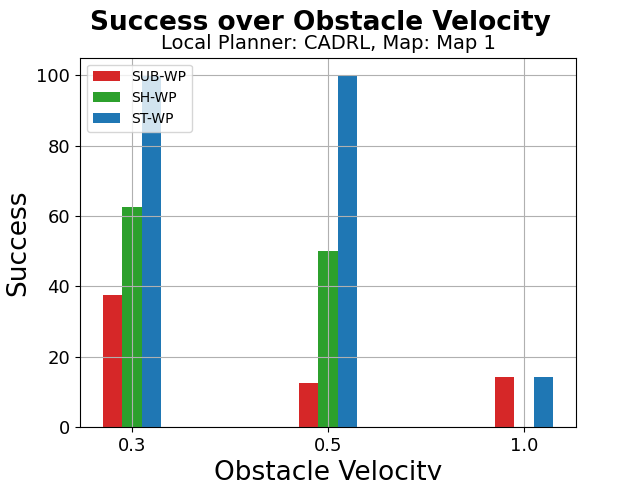}
		
		\label{fig:1}
	\end{subfigure}\hfil % <-- added
	    \begin{subfigure}{0.25\textwidth}(b)
		 \centering % <-- added
		\includegraphics[width=\linewidth]{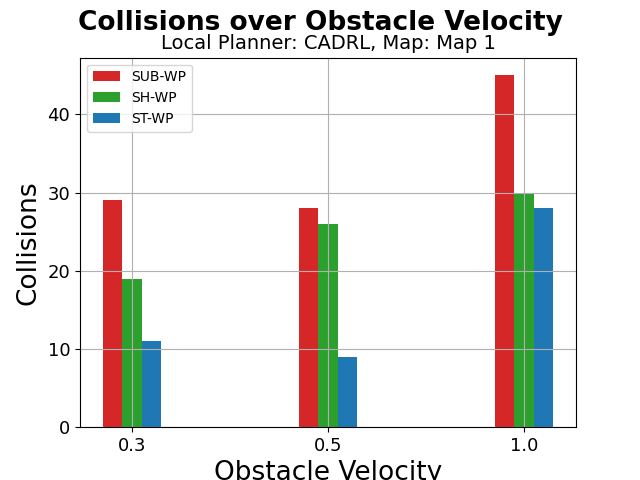}
		
		\label{fig:1}
	\end{subfigure}\hfil % <-- added
	\begin{subfigure}{0.25\textwidth }(c)
	 \centering % <-- added
		\includegraphics[width=\linewidth]{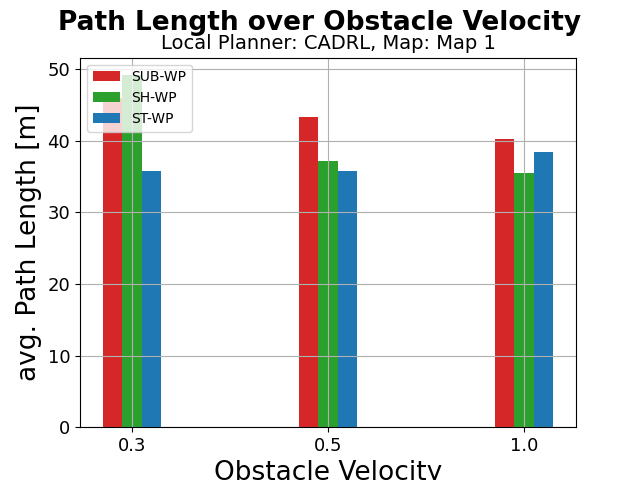}
		
		\label{fig:2}
	\end{subfigure}\hfil % <-- added
	\begin{subfigure}{0.25\textwidth}(d)
	 \centering % <-- added
		\includegraphics[width=\linewidth]{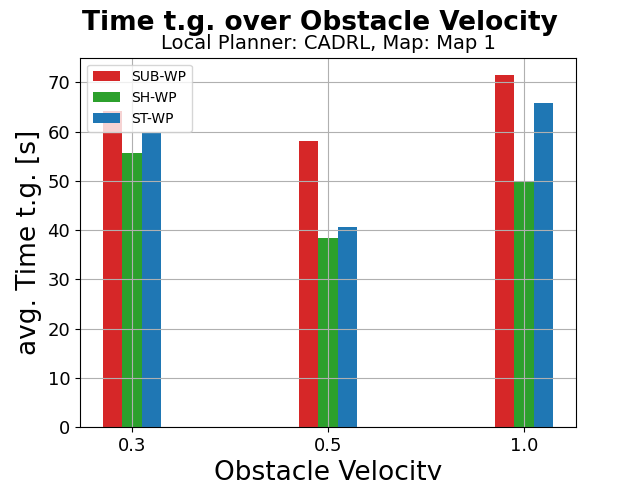}
		
		\label{fig:3}
	\end{subfigure}
	
	\caption{Upper Row: Trajectories of all planners on the test scenario with an obstacle velocity $v_{obs}=0.5 m/s$. (a) 5 dynamic obstacles , (b) 10 dynamic obstacles, (c) 20 dynamic obstacles. The colored dots indicate a collisions. The colored circles indicate a large number of collisions. The trajectory of the obstacles are black. Lower row: Quantitative comparison between the three approaches. (a) Success rate, (b) Collisions (c) Path length, (d) Time to reach goal over the obstacle velocity. }
	\label{quali}
\end{figure*}

\section{Results and Evaluation}
To evaluate our space-timed waypoint generator, we compared it against two baseline approaches. The first one is a simple subsampling of the global plan, which we denote as SUB-WP. The second one is an approach of our previous work \cite{kastner2021towards}, which spawns a waypoint depending on the robot's position. It is denoted as SH-WP. The approach presented in this paper is denoted as ST-WP. As a local planner, CADRL \cite{everett2018motion}, a DRL-based obstacle avoidance approach is utilized to follow subgoals computed by the waypoint generators.
The experiments were conducted on an office map with static obstacles and an increasing number of dynamic obstacles. We assumed static obstacles are known to the robot on the given global map. The dynamic obstacles were unknown to the robot and could only be sensed locally within the sensor range, which is 3.5 $m$ in our robot setup. The obstacles move with a back and forth movement between two positions with predefined lengths. The obstacles are triggered only when the robot is nearby. Thus, the robot may meet obstacles that suddenly appear in front of it, which may add difficulties to the obstacle avoidance task. If the distance between the robot and any static or dynamic obstacle is less than 0.35 $m$, a collision is published and counted. Each test trial continues until the robot reaches the goal position or a timeout of 3 min is exceeded. When the robot has reached this goal, the scenario will be reset by the task generator. We define a test trial as successful if the robot reaches the goal within the time limit and with less than 2 collisions.

\subsection{Qualitative Results}
Figure \ref{quali} illustrates the robot's trajectories and collision zones over different numbers of obstacles with a velocity of 0.3 m/s. As observed, the higher the obstacle speed, the higher the collision rates for all approaches. However, the SH-WP and SUB-WP result in a significantly higher number of collisions compared to the ST-WP. Furthermore, it is observed that the ST-WP maintains a robust and efficient path to the goal, whereas the other two approaches contain roundabout paths especially starting from the position [x: 11,y: 11]. Even with increasing obstacle velocity, the trajectories of our proposed ST-WP approach maintain their robustness. This is due to the fact that the approach considers the obstacle's position and velocity at every time step and thus can provide a safer waypoint. The following quantitative evaluation further outlines these findings.

\subsection{Quantitative Results}
The quantitative results of the robot performance with different subgoal modes on the office map over different obstacle velocities and fixed number of obstacles ($N_{obs}=20$) are shown in Fig. \ref{quali}.
For the scenarios with 20 obstacles and a velocity of 0.3 and 0.5 m/s, our ST-WP approach accomplishes a 100 percent success rate compared to over 60 percent for the SH-WP and over 35 percent for the SUB-WP. For a velocity of 0.5 m/s, the success rate for SH-WP and SUB-WP drops to over 50 and over 10 percent respectively, while our ST-WP maintains a 100 percent success rate. However, for an obstacle velocity of 1 m/s, the success rate drops rapidly for all approaches to under 15 percent. Our proposed ST-WP achieves the lowest collision rates with only 11 collisions for 0.3 m/s and under 10 for 0.5 m/s, while the other two approaches reach over 25 collisions. These results demonstrate the superiority of our proposed waypoint generator in terms of safety and robustness compared to the other two baseline approaches.
In terms of efficiency, all approaches perform similarly with a slightly more efficient performance by our ST-WP approach.
The subsampling method proves to be the worst in terms of efficiency because the robot has to traverse all subgoals along the initial global trajectory before reaching the final goal. If subsampled waypoints are blocked by dynamic obstacles, the robot needs to take extra effort to reach that waypoint before approaching the next one.
\subsection{Discussion}
The results demonstrate the superiority of our ST-WP approach. It causes fewer collisions while being slightly more efficient than the baseline approaches SUB-WP and SH-WP. This is due to the fact that our approach considers obstacle positions and velocities to calculate an efficient path based on which a subgoal is selected that guides the robot within a highly dynamic environment. Due to the usage of efficient planning methods, the calculations can be done efficiently and quickly. The paths are smoother due to the refinement step incorporating the EGO trajectory optimizer.
Another important aspect is that the safety and efficiency performance of SH-WP and SUB-WP is highly dependent on the quality of the global path because it is the only source to calculate the waypoints. Since the global path is not updated until specific conditions are met, the quality of the future part of the path is not ensured and may lead to worse results in certain scenarios where the path is in an obstacle-dense area. Moreover, in a large-scale environment, frequent global replan consumes much more time. By using the landmark method, the cost for global replan is reduced.

%% file: ieeeconf/content/5-conclusion.tex
\section{Conclusion}
In this paper, we proposed a hierarchical navigation approach combining model-based optimization for the mid planner with a DRL-based local planner for long-range navigation in highly dynamic environments. Our approach efficiently searches the state-time space utilizing a timed Delaunay Trianagle graph to encode dynamic obstacles and efficiently generate a collision free trajectory using a modified timed A-Star approach. The trajectory is further refined by an EGO optimizer and an optimized subgoal is generated. The resultant waypoint from the mid planner has incorporated both global information and dynamic obstacles within the sensor range.
Subsequently, we evaluated our approach against two baseline approaches and found an enhanced performance in terms of navigational safety and robustness. Future work include the incorporation of semantic information into the navigation system, the combination with other local planners as well as a DRL-based training of the joint system. Furthermore, we aspire to evaluate the approach in real environments and robotic hardware.

% \begin{table}[H]
% 	\setlength{\tabcolsep}{2pt}
% 	\renewcommand{\arraystretch}{0.7}
% 	\centering
% 		\caption{Rewards and Penalties}
% 	\begin{tabular}{lllll}
% 		\hline
% 		Event  & Value   & Description    \\ \hline
% 		Goal    & 100        & Agent reaches goal          \\ 
% 		Towards goal    &0.1       &Agent moves towards goal        \\ 
% 		Col-Human &-100        & Agent collides with human          \\ 
% 		Col-Robot &-80        & Agent collides with robot          \\ 
% 		Away from goal    & -0.2       & Agent moves away from goal        \\ 
% 	    Hit Static Object    & -10      & Agent hits static obstacle        \\ 
% 		\hline
% 	\end{tabular}

% 	\label{actions}
% \end{table}